 \documentclass[letterpaper, 10pt, conference]{ieeeconf}
 \IEEEoverridecommandlockouts
\usepackage[cmex10]{amsmath}
\usepackage{amsmath}
\usepackage{amssymb}
\usepackage{amsfonts}
\usepackage{amsthm}
\usepackage{algorithm}
\usepackage{algpseudocode}
\usepackage{graphicx}
\usepackage{dblfloatfix}
\usepackage{subcaption}
\usepackage{epsfig}
\usepackage{cite}
\usepackage{tensor}
\usepackage{comment}
\usepackage{color}

\usepackage[dvipsnames]{xcolor}
\definecolor{rephrase}{RGB}{219, 10, 10}

\usepackage{diagbox}
\usepackage{gensymb}
\usepackage{environ}         
\usepackage{etoolbox}        
 
\graphicspath{{figures/}}
\DeclareGraphicsExtensions{.eps}
\theoremstyle{definition}

\usepackage{tablefootnote}
\newlength{\myl}
\let\origequation=\equation
\let\origendequation=\endequation

\RenewEnviron{equation}{
  \settowidth{\myl}{$\BODY$}                       
  \origequation
  \ifdimcomp{\the\linewidth}{>}{\the\myl}
  {\ensuremath{\BODY}}                             
  {\resizebox{\linewidth}{!}{\ensuremath{\BODY}}}  
  \origendequation
}

\begin{document}

\title{\LARGE \bf Vision-Based Control for Robots by a Fully Spiking Neural System Relying on Cerebellar Predictive Learning}

\author{Omar Zahra, David Navarro-Alarcon and Silvia Tolu%
\thanks{This work is supported in part by the Research Grants Council (RGC) of Hong Kong under grant number 14203917.}%
\thanks{O. Zahra and D. Navarro-Alarcon are with The Hong Kong Polytechnic University, Department of Mechanical Engineering, Kowloon, Hong Kong. Corresponding author e-mail: {\texttt{\small dna@ieee.org}}.}
\thanks{S. Tolu is with Technical University of Denmark, Department of Electrical Engineering, Copenhagen, Denmark.%
}}

\bstctlcite{IEEEexample:BSTcontrol}

\maketitle
\thispagestyle{empty}
\pagestyle{empty}

\begin{abstract}
The cerebellum plays a distinctive role within our motor control system to achieve fine and coordinated motions. While cerebellar lesions do not lead to a complete loss of motor functions, both action and perception are severally impacted. Hence, it is assumed that the cerebellum uses an internal forward model to provide anticipatory signals by learning from the error in sensory states. In some studies, it was demonstrated that the learning process relies on the joint-space error. However, this may not exist. This work proposes a novel fully spiking neural system that relies on a forward predictive learning by means of a cellular cerebellar model. The forward model is learnt thanks to the sensory feedback in task-space and it acts as a Smith predictor. The latter predicts sensory corrections in input to a differential mapping spiking neural network during a visual servoing task of a robot arm manipulator. In this paper, we promote the developed control system to achieve more accurate target reaching actions and reduce the motion execution time for the robotic reaching tasks thanks to the cerebellar predictive capabilities.
\end{abstract}


\section{INTRODUCTION}

Our hierarchical motor control system is composed of several areas that are responsible for different functions \cite{merel2019hierarchical}. The high-level areas are in charge of decision making and planning sequence of motion. On the other hand, low-level areas control muscles and regulate both forces and velocities while interacting with dynamic environments. The motor cortex transforms the motion intentions into motor commands, then sends them to the brainstem and the cerebellum. Research studies have shown that patients with cerebellar lesions suffer from clumsy staggering movements \cite{kandel2000principles}.

Hence, various computational models were built around the cerebellar theories on internal model formation for coordination, precision, and accurate timing \cite{porrill2013adaptive}. In particular, the cerebellum seems to act as an inverse model to generate corrective motor commands, as a forward model to enhance sensory predictions, and as a combination of both models \cite{wolpert1998multiple, corchado2019integration}. Considering the effects of cerebellar lesions on both action and perception, it was suggested that the cerebellum is particularly involved in predicting the next sensory states and that the learning process is driven by the error in sensory states \cite{bastian2011moving}. In this paper, we take advantage of the cerebellum capabilities to process the sensory information and predict sensory states to enhance robotic target reaching actions.

Most of the studies consider the cerebellar learning relying on error in joint space \cite{antonietti2019, abadia2019robot, tolu2020cerebellum, Capoleietal2020}. However, in many cases a reference signal in joint space may not exist and thus, learning can be only be carried out based on task-space error.
Cerebellar models were categorized in \cite{van2016neurorobotics}, which they can be either a state-encoder-driven model, functional or cellular model. More attention is given to functional models to approximate functions executed by each cerebellar layer, which makes it more appealing from computer-science perspective \cite{tolu2020cerebellum}. While the cellular models are more challenging, it is not crucial only to verify and develop theories about the learning mechanisms, but to gain insight as well on how neurons act on individual and population basis to achieve such task \cite{abadia2019robot}. Additionally, the spiking nature of neurons makes it possible to run the model on neuromorphic chips, thus allowing less energy consumption and real-time computing \cite{furber2016large, Bairaetal2017}.

In \cite{Capoleietal2020}, the introduced model is based on the adaptive filter theory. The developed control scheme combines machine learning and some techniques from computational neuroscience for the sensorimotor optimization. However, it does not model the spiking nature of the cerebellum. 
In \cite{abadia2019robot}, a compliant controller was built based on a spiking cerebellar model. The cerebellar model relies on joint-space errors based on reference values from a trajectory generator. Thus, it lacks the ability to directly learn from sensory feedback.

In this work, we developed a cellular-level cerebellar controller that guides the motion of a robot based on real-time sensory feedback information. First, a sensorimotor differential map is coarsely built and trained through a motor babbling process. Next, a forward cerebellar internal model is formed to predict and adjust the discrepancies in sensory readings to minimize the deviation from reference targets.

For the first time, a control system that integrates two spiking neural networks (SNNs) to, respectively, emulate the cerebellum's role in anticipatory and adaptive control, and form the differential sensory maps during a vision-guided motion task for a physical robot, is presented. The SNNs incorporate most of the biological neuronal dynamics which include complex and realistic firing patterns based on spikes \cite{maass1997networks}.
Besides, the novelty of the work is that the forward cerebellar predictions are integrated as a Smith predictor \cite{tolu2020cerebellum} and they are supervised by the error in task space.

To show the effectiveness of our approach, we present a detailed experimental study in which  a robot arm performed vision-guided manipulation tasks like reaching a target and following a contour. Results show that the deviation from the desired path is reduced and the execution time is decreased thanks to the cerebellar sensory predictions, which is coherent with experimental results from humans \cite{dewolf2016spiking}.

In the following sections we are describing the developed SNNs and controller (Sec. II), the experimental setup and results (Sec. III), and finally, we are analysing the results and main conclusions (Sec. IV).

\section{METHODS}\label{sec:methods}

A biologically inspired control system shown in Fig.\ref{fig:chart} for voluntary movements is built by integrating two spiking computational models: the differential map ($DM$) and the cerebellar-like controller ($CB$). The sensorimotor map in $DM$ block is formed thanks to the motor babbling process and it receives the modulated sensory input by the $CB$ outcomes. 

\subsection{Cerebellar-Based Smith Predictor}

\begin{figure}[b]
\centering
\includegraphics[width=0.6\linewidth]{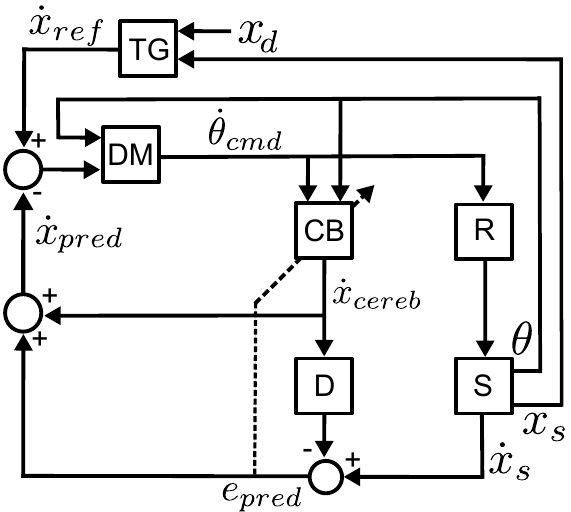}
\caption{The block diagram of the control system. The cerebellum $CB$ acts as a predictor for the next robot state based on $\dot{\theta}_{cmd}$ generated by the differential map $DM$. The sensors $S$ provide the current state of the robot $R$ with a delay $D$, which is compared to the desired state by the Target Generator block $TG$ to generate $\dot{x}_{ref}$.}
\label{fig:chart}
\end{figure}

The cerebellar computational model proposed in this study forms a forward model to predict the next sensory states based on the current robot states and the desired spatial velocity. The forward model allows the designed controller to act as a Smith predictor, which is known to be capable of compensating long time delays (see Fig.\ref{fig:chart}). A dead time is introduced in a control system due to the time needed for sensing, processing of the inputs, computing the control output and actuation, similar to that in a biological system for the sensory signals to travel through the nervous system for processing and generating an adequate motor command which reaches the muscles. In our control system, the dead time ($D$) is introduced by the delay in sensor readings ($S$) and the time needed for the robot ($R$) to take action.  

The $DM$ is the differential map \cite{zahra2020differential} needed to generate motor commands in manner similar to the motor cortex, thus acting as an inverse dynamic model, where motor commands ($\dot{\theta}_{cmd}$) are generated based on the current states (the spatial velocity $\dot{x}_s$ and joint angles $\theta$). In robotics, this is usually expressed in the form of an inverse Jacobian matrix $J_{DM}^{\#}(\theta)$ which estimates the required motor command to obtain a target spatial velocity at the end effector $\dot{x}_{ref}$. While considering the delay $\tau$ mentioned earlier it can be expressed as: $\dot{\theta}(t) = J_{DM}^{\#}(\theta)\dot{x}_s(t-\tau)$ given that no correction is applied to make up for the delay introduced. The $CB$ block acts as a forward model to predict the spatial velocity after applying the generated motor commands:

\begin{equation}
    \dot{x}_{cereb}(t) = \tilde{J}(\theta)\dot{\theta}(t)
\end{equation}

where $\dot{x}_s(t-\tau)$ refers to the delayed sensory readings for the actual cartesian velocity of the end effector, and $ \dot{x}_{cereb}(t)$ is the estimated cartesian velocity based on the command $\dot{\theta}(t)$. In case of an optimal controller, the cerebellar-like model would perfectly resemble the forward model of the robot (ie. $\tilde{J}(\theta) = J(\theta)$), where $J(\theta)$ is the Jacobian matrix that describes the real robot; in the studied case, our cerebellar-like model acts as a predictor to correct the differential map estimations. Thus, $CB$ outcomes approximate $J_{DM}$ itself. The difference between the predicted and actual spatial velocity is used for training the cerebellar model, so it minimizes such difference, such that finally $\tilde{J(\theta)} = J_{DM}(\theta)$ when $\dot{x}_{cereb}=\dot{x}_s$ at all the points in the studied space. The estimations can then be interpreted as:
\begin{align}
\begin{split}
     \dot{x}_{cereb}(t-\tau) &= \tilde{J}(\theta)\dot{\theta}     \\
    &= J_{DM}(\theta)(J_{DM}^{\#}(\theta)\dot{x}_s(t-\tau)) \\
    &= \dot{x}_s(t-\tau)
\end{split}
\end{align}

Then, the $CB$ predicts the next sensory state of the robot ($\dot{x}_{cereb}$) which is compared to the corresponding sensory readings ($\dot{x}_{s}$) in the next cycle. The discrepancy in the prediction ($e_{pred}$) is used to train the $CB$ to give better predictions in future cycles:
 \begin{equation}
     e_{pred} = \dot{x}_s - \dot{x}_{cereb}(t-\tau)
 \end{equation}
 Then, $e_{pred}$ is added to the current cycle prediction to make up for the expected error in current readings to obtain $\dot{x}_{pred}$:
 \begin{equation}
     \dot{x}_{pred} = \dot{x}_{cereb}(t) + e_{pred}
 \end{equation}
 The difference between the $\dot{x}_{ref}$ desired spatial velocity and $\dot{x}_{pred}$ shall provide the necessary correction to make up for the error in the approximations of the differential map ($DM$). The $\dot{x}_{ref}$ is updated by the Target Generator block ($TG$) every cycle by comparing the current position $x_s$ to the target position $x_d$:
 \begin{equation}
     \dot{x}_{ref} = \frac{x_d-x_s}{\parallel x_d-x_s\parallel}
 \end{equation}

\subsection{The Differential Mapping Spiking Neural Network}

\begin{figure}[!t]
\centering
\begin{subfigure}{0.5\columnwidth}
\centering
\includegraphics[width=\columnwidth]{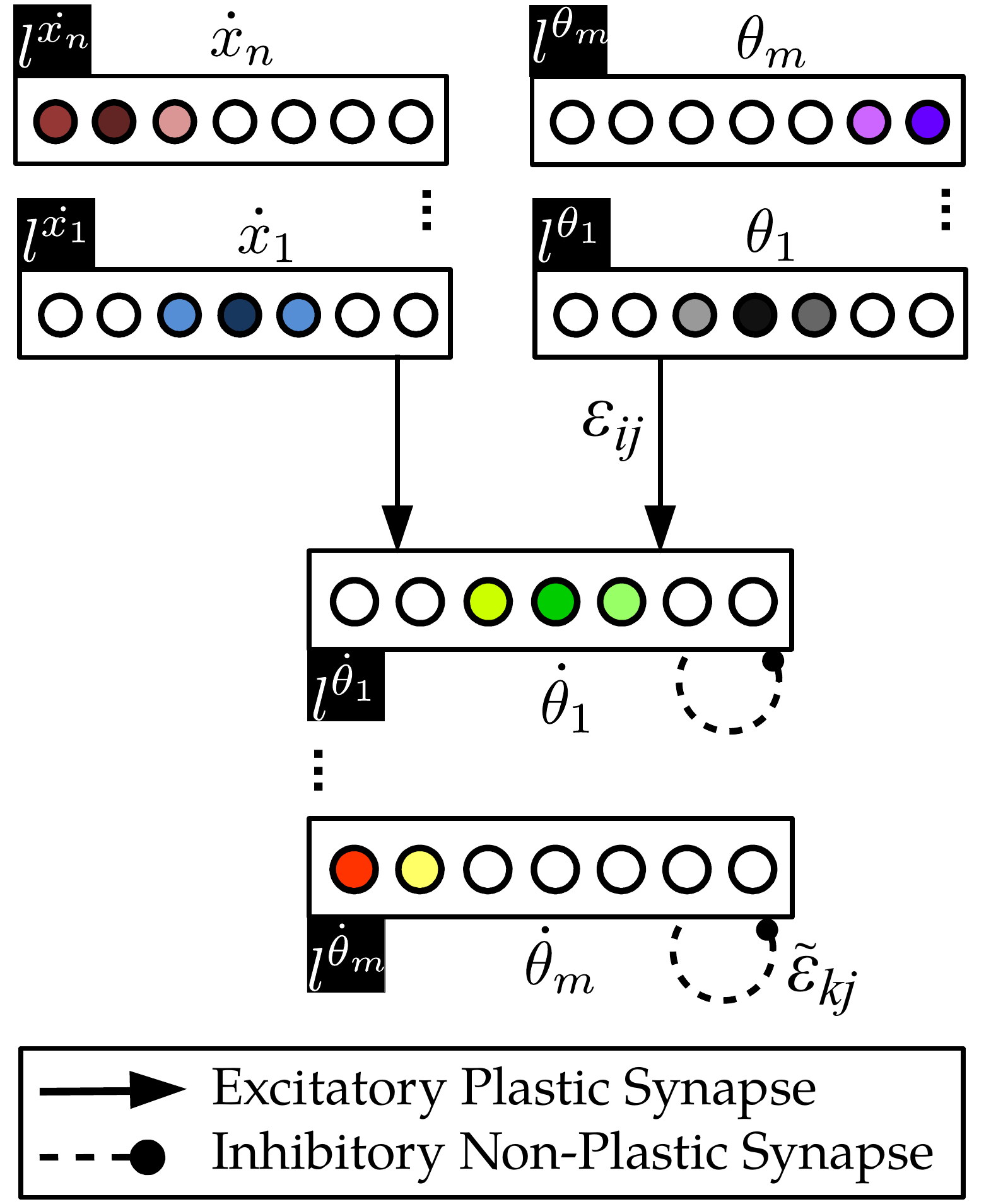}
\caption{}
\label{fig:net}
\end{subfigure}%
\begin{subfigure}{0.5\columnwidth}
\centering
\includegraphics[width=1.0\linewidth]{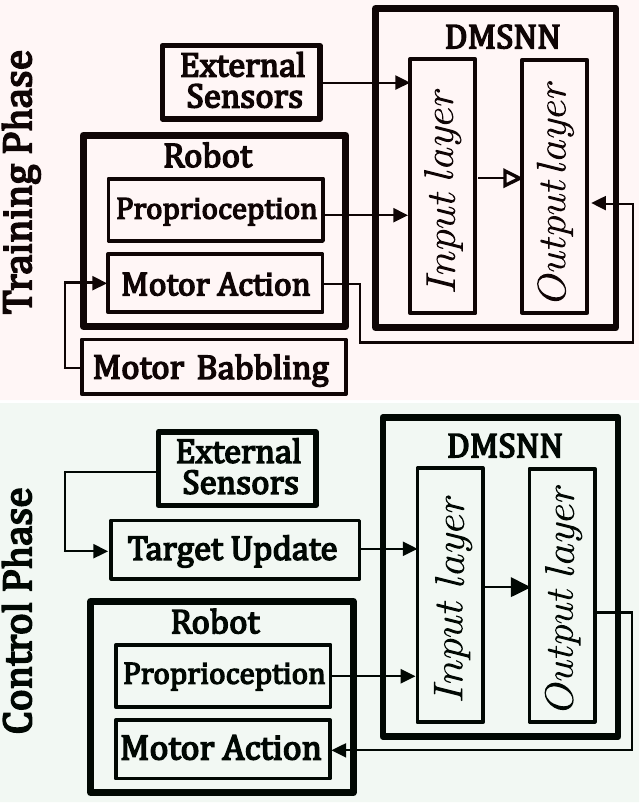} 
\caption{}
\label{fig:train_test_breif}
\end{subfigure}
\caption{(a) The layout of the $DM$ block. (b) A schematic diagram for both the training and control phases for the SNN.}
\label{fig:dmsnn}
\end{figure}

A SNN is composed of two layers (an input and output layer) that are connected through all to all plastic synapses to map two correlated spaces and represent the transformation between these spaces. As the map correlates in this study the velocities in joint space -as an output- and in task space -as an input-, thus it acts as a differential map (block $DM$). For a manipulator of $n$ and $m$ degrees of freedom (DOF) in task space and joint space, $n$and $m$ assemblies of neurons are needed, respectively, each encoding one corresponding dimension of the relative space. Hence, the input layer consists of $n+m$ assemblies; $n$ assemblies ($l^{\dot{x}_{1:n}}$) to encode the spatial velocity ($\Dot{x}\in\mathbb R^n$) and $m$ assemblies ($l^{\theta_{1:m}}$) to encode the angular positions of the joints ($\theta\in\mathbb R^m$), while the output layer consists of $m$ assemblies ($l^{\dot{\theta}_{1:m}}$) to encode the angular velocities of the joints ($\Dot{\theta}\in\mathbb R^m$) as shown in \ref{fig:net}. As in this case each dimension of task space is connected to each dimension of joint space, the network resembles a Jacobian-like transformation where $\Dot{x}=J(\theta)\Dot{\theta}$, such that $J_{ij}=\partial x_{i}/\partial \theta_{j}$,
where $J(\theta)\in\mathbb R^{n\times m}$ is the Jacobian matrix.
The input and output neurons are connected through all to all plastic excitatory and inhibitory synapses. Local inter-inhibitory synapses are added within each assembly of the output layer with increasing inhibition from one neuron to distal neurons.
The plastic synapses are modulated during the training phase as indicated in Fig.\ref{fig:train_test_breif}. The internal sensors (proprioception), such as motor encoders, and external sensors (exterioception), such as cameras, provide the data, collected through motor babbling, needed to be introduced to both the input and output layer during the training phase. The encoding of variables is based on the central (preferred) value $\psi_{c}$ assigned to each neuron, with the contribution of the whole assembly utilizing ``population coding'' \cite{amari2003handbook}).

The synapses are modulated accordingly to form the differential map, and the SNN in $DM$ can be used to guide the robot in servoing tasks after several iterations \cite{zahra2020differential}. However, the SNN in $DM$ represents a coarse map which lacks in accuracy and precision and needs to be modulated to be adequate for fine motion control.

\subsection{The Proposed Forward Cerebellar Model}

The cerebellar micro-circuit consists of the granular layer, Purkinje layer, and molecular layer. The granular layer contains the granule cells (GC), which make up most of the cells in the brain (around 80\% of the total number of neurons in the brain \cite{herculano2009human}), along with Golgi cells and the mossy fiber ($MF$) axons. The Purkinje layer contains the Purkinje cells ($PC$), which are considered to be the key player with it's distinctive firing pattern for different inputs. The molecular layer contains the parallel fibers ($PF$) intersecting with climbing fibers ($CF$) (carrying signals from inferior olive) and axons of $PC$, along with basket and stellate cells.

In this work, we modeled the cerebellar microcircuit to achieve fine motor control as it provides corrections for the sensory readings introduced to the differential map developed ($DM$), which is somewhat functionally similar to the cerebellum-motor cortex loop \cite{ishikawa2016cerebro}. 
Based on the forward model theory \cite{tanaka2019neural}, the $MF$ encode both the sensory inputs and motor commands, which is the output of the $DM$, where each dimension of either the joint space or task space is encoded by an assembly of neurons. The connections between the $MF$ and the granular layer ($GC$) allows for a sparse coding of the robot states. The inferior olive ($IO$) introduces the teaching signal in terms of the task-space error through the $CF$, calculated as the difference between the predicted and the actual spatial velocities, in this case. By using the plasticity of synapses between $GC$ and $PC$, these connections are modulated under supervision of teaching signals from $CF$ to encode the desired correction at the different states of the robot. The $PC$ the inhibits the corresponding Deep Cerebellar Nuclei ($DCN$) to allow for the right value to be decoded from the activity of these neurons. 

The $MF$ is divided into two groups, one for encoding values in joint space ($MF_{JS}$) and the other for encoding values in task space ($MF_{TS}$), having total number of assemblies in the whole $MF$ layer equal to $J_{MF}$. The $PC$, $IO$ and $DCN$ consist of $n_{TS}$ groups of neurons and each group has two assemblies of neurons, one for positive and one for negative change of the corresponding degree of freedom (DOF).

\begin{figure}[!b]
\centering
\includegraphics[width=1.0\linewidth]{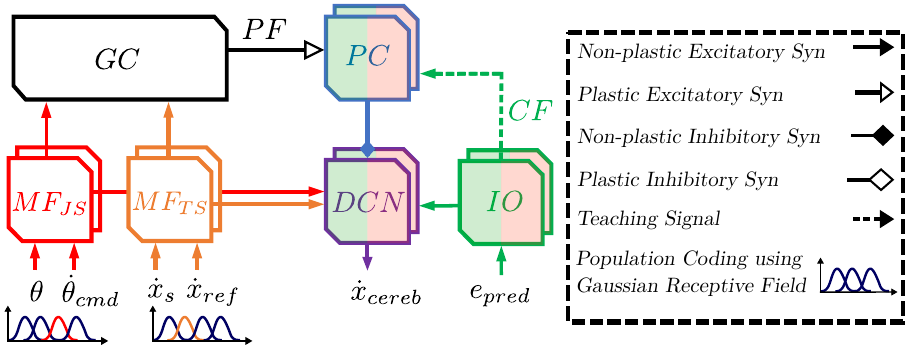} 
\caption{(a) A schematic diagram for the cerebellum. The $PCs$, $IOs$ and $DCNs$ are divided into two to represent positive and negative changes in the corresponding DOF. }
\label{fig:cereb}
\end{figure} 

\begin{table}[!b]
\caption{$CB$ Network Parameters}

\begin{tabular}{ |p{2.2cm}||p{0.75cm}|p{0.75cm}|p{0.75cm}|p{0.75cm}|p{0.75cm}|  }
 \hline
 \multicolumn{6}{|c|}{Properties of neurons} \\
 \hline
 Area & $a$ & $b$ & $c$ & $d$ & $\mathcal{N}$ \\
 \hline
 $MF$ & 0.1 & 0.2 & -65. &2. & 20\\
 $GC$ & 0.02 & 0.25 & -65. &2. & 1000\\
 $PC$ & 1. & 1.5 & -60. &0. & 8\\
 $IO$ & 0.1 & 0.2 & -65. &2. & 8\\
 $DCN$ & 0.05 & 0.1 & -65. &2. & 4\\
 \hline
\end{tabular}

\begin{tabular}{ |p{2.2cm}||p{1.12cm}|p{1.02cm}|p{1.02cm}|p{1.02cm}|  }
 \hline
 \multicolumn{5}{|c|}{Properties of synapses} \\
 \hline
 Projection & Type & P & $\varepsilon_{init}$ & $\varepsilon_{max}$\\
 \hline
 $MF\rightarrow GC$ & Rand & 4 & 1.6 &-\\
 $MF\rightarrow DCN$ & A2A & 1.0 & 1.0 &-\\
 $GC\rightarrow PC$ & Prob & 0.8 & 0.2 &8.0\\
 $IO\rightarrow PC$ & O2O & - & 500.0 &-\\
 $IO\rightarrow DCN$ & O2O & - & 1.7 &-\\
 $PC\rightarrow DCN$ & O2O & - & -5.0 &-\\
 \hline

\end{tabular}
 \footnotetext  'The Type column defines the pattern of connections from layer A to B  in the 'Projection' column with a parameter value 'P'. 'Rand' means that a P random neurons from A connect to one neuron in B. 'Prob' means that there is a probability P for each neuron in A to connect to each neuron in B. 'A2A' means All-to-All connections, and all neurons from A are connected to All neurons in B. 'O2O' means One-to-One connections, and each neuron A is connected to one corresponding neuron in B.
\label{table:cb_params}
\end{table}

The training of the cerebellum starts after the SNN in $DM$ is trained for enough iterations to form a coarse map of the transformation between the task space and motor space.

The input to the $MF$ is encoded based on population coding using Gaussian receptive fields, and the input current to the $i^{th}$ neuron in the $j^{th}$ assembly of MF $\alpha_{MF_{i,j}}(t)$ can be calculated from the following formula:
\begin{equation}
\alpha_{MF_{i,j}}(t) = \exp {\left(\dfrac{-\Vert{\psi_{MF_j} -\psi_{MF_{i,j}}  \Vert^{2}}}{2\sigma_{MF_j}^{2}}\right)}
\label{eq:f_MF}
\end{equation}
where $\psi_{MF_j}$ is the input value to the $j^{th}$ assembly of MF, and $\sigma_{MF_j}$ is calculated based on number of neurons per layer $N_{l}$, and the range of change of the variable to be encoded from $\Psi_{MF_{j_{min}}}$ to $\Psi_{MF_{j_{max}}}$. $\psi_{MF_{i,j}}$ is the central value of the $i^{th}$ neuron in the $j^{th}$ assembly of $MF$.

Each neuron in the $GC$ is connected to one random neuron from each assembly of neurons in the $MF$ layer. The $MF$ connects, as well, through excitatory synapses to the $DCN$ through all to all connections. The $GC$ then projects to the $PC$ through plastic excitatory synapses with a probability $P_{GC-PC}$, and these connections are the $PF$. The parameters are set such that $PC$ synapses are only modulated when neurons in $IO$ are active. The connection between $IO$ and $PC$, is given by $CF$, which are one to one synapses such that neurons of the same group (ie. same DOF) and same direction (ie. positive or negative change) are connected to each other, to ensure only the right $GC-PC$ synapses are modulated. The $IO$ is connected as well through excitatory synapses to $DCN$. The activity of $IO$ neurons can be given by:
\begin{align} 
        \alpha_{IO+_j} = 
        &\left\{
            \begin{array}{ll} 
          \alpha_{IO_{max}} & e_{pred} > 0 \\
          0 & e_{pred}\leq 0
       \end{array}
        \right.  
        \\
        \alpha_{IO-_j} = 
        &\left\{
            \begin{array}{ll} 
          0 & e_{pred}\ge 0 \\
          \alpha_{IO_{max}} & e_{pred} < 0
        \end{array}
        \right.
\end{align}

where $\alpha_{IO+_j}$ and $\alpha_{IO-_j}$ are the average firing rates of the positive and negative IO assemblies of neurons of the $j^{th}$ DOF, respectively, while $\alpha_{IO_{max}}$ is the maximum firing rate of each neuron in the $IO$. 

Similarly, the connections between $PC$ and $DCN$ follow the same concept to enable the activation of the right group of neurons. The output from the $DCN$ is then decoded based on the activation of neurons such that:

\begin{equation}
    \dot{x}_{cereb_j} = \frac {\Sigma \alpha_{DCN+_{i,j}} - \Sigma \alpha_{DCN-_{i,j}}}{{\alpha_{DCN_{max}}} * n_{DCN}} *\dot{x}_{max_j}
\end{equation}

where $\dot{x}_{cereb_j}$ is the predicted velocity of the $j^{th}$ DOF, $\alpha_{DCN+_i,j}$ and $\alpha_{DCN-_i,j}$ are the firing rates of the $i_{th}$ neuron in the positive and negative assemblies of the $j^{th}$ DOF, respectively, $n_{DCN}$ the number of DCN neurons in each assembly and $\alpha_{DCN_{max}}$ is the maximum firing rate of each neuron in the $DCN$. 

The network parameters are set to allow the $PF$ synapses to change with an adequate pace and reach a certain value after several iterations, such that enough training data is provided. The final synaptic strength of the plastic connections enables $PC$-$DCN$ inhibitions at the proper timing thanks to the teaching signals provided earlier from the $IO$. The neurons in the $PC$ are chosen to act as bistable neurons such that high firing rates can be obtained, yet the firing action can still be controlled by the teaching signal to induce synaptic gating. For this, an inhibiting signal is introduced all the time to the $PC$ as a substitute of the inhibition introduced in the biological model from basket and stellate cells. The strength of the synaptic connections between $IO$ and $PC$ ($CF$) are set to a big value, analogous to having around 300 synapses from each $CF$ to each $PC$ \cite{van2016neurorobotics} to make sure that each spike in $IO$ is enough to evoke spikes in the $PC$, but yet having one synapse instead to save computational power required to represent redundant signals.

\subsection{Neuron Model}

Izhikevich's simple model of spiking neurons \cite{izhikevich2003simple} is adopted for its capability of reproducing various firing patterns while holding a balance between computational cost and biological plausibility \cite{izhikevich2003simple}.
The model is formulated by a set of differential equations:
\begin{align}
\dot{v} & = f(v,u) = 0.04v^2+5v+140-u+I    \label{eq:update_v} \\
\dot{u} & = g(v,u) = a(bv-u)               \label{eq:update_u}
\end{align}
The after-spike membrane potential is reset based on:
\begin{equation}
\label{eq:reset}
\text{if }v\ge30 \text{ mV}, \quad
\text{then } v\leftarrow c,~ u\leftarrow (u+d)
\end{equation}
where $v$ is the membrane potential and $u$ is the membrane recovery variable.
The parameter $a$ determines the decay rate (time constant) for $u$, $b$ controls the sensitivity of $v$ to the sub-threshold fluctuations, 
$c$ describes the after-spike membrane potential, and $d$ gives the after-spike value of $u$. 
The term $I$ gives the summation of external currents.

\subsection{Synaptic Connections}
The adopted learning rule is the Spike-timing dependent plasticity (STDP), where increase or decrease in the strength of the connections is dependent upon the relative timing of spikes in pre-synaptic and post-synaptic neurons \cite{stdp}.

The anti-symmetric STDP learning rule \cite{stdp_asym_pic} is applied for the plastic synapses in the cerebellum model, and can be expressed as:
\begin{equation}
\Delta {\varepsilon}_{ij} = \left\{
        \begin{array}{ll}
            -S_{a} \exp \left( {-\Delta t}/{\tau_{a}} \right) & \quad \Delta t \leq 0 \\
            \\
             S_{b} \exp \left( {-\Delta t}/{\tau_{b}} \right) & \quad \Delta t > 0
        \end{array}
    \right.  
\label{eq:asym_STDP}
\end{equation}
where $S_{a}$ and $S_{b}$ determine the magnitude of the depression and potentiation of synaptic weights, respectively, while $\tau_{a}$ and $\tau_{b}$ decide the time window for occurrence of depression and potentiation, respectively. 

For the $DM$, the symmetric model \cite{woodin2003coincident} is adopted, and can be expressed as:
\begin{equation}
\Delta {\varepsilon}_{ij} = S \left( 1-\left( {\Delta t}/{\tau_{1}} \right)^2\right) \exp\left({|\Delta t|}/{\tau_{2} } \right)
\label{eq:sym_STDP}
\end{equation}
where $S$ decides the magnitude of the change in synaptic weights, the ratio between the two variables $\tau_{1}$ and $\tau_{2}$ controls the time window through which potentiation and depression occurs, and $\Delta t$ is the timing difference of spikes at post-synaptic $t_{post}$ and pre-synaptic $t_{pre}$ neurons.

\begin{figure}[b]
\centering
\begin{subfigure}{0.49\columnwidth}
\includegraphics[width=\linewidth]{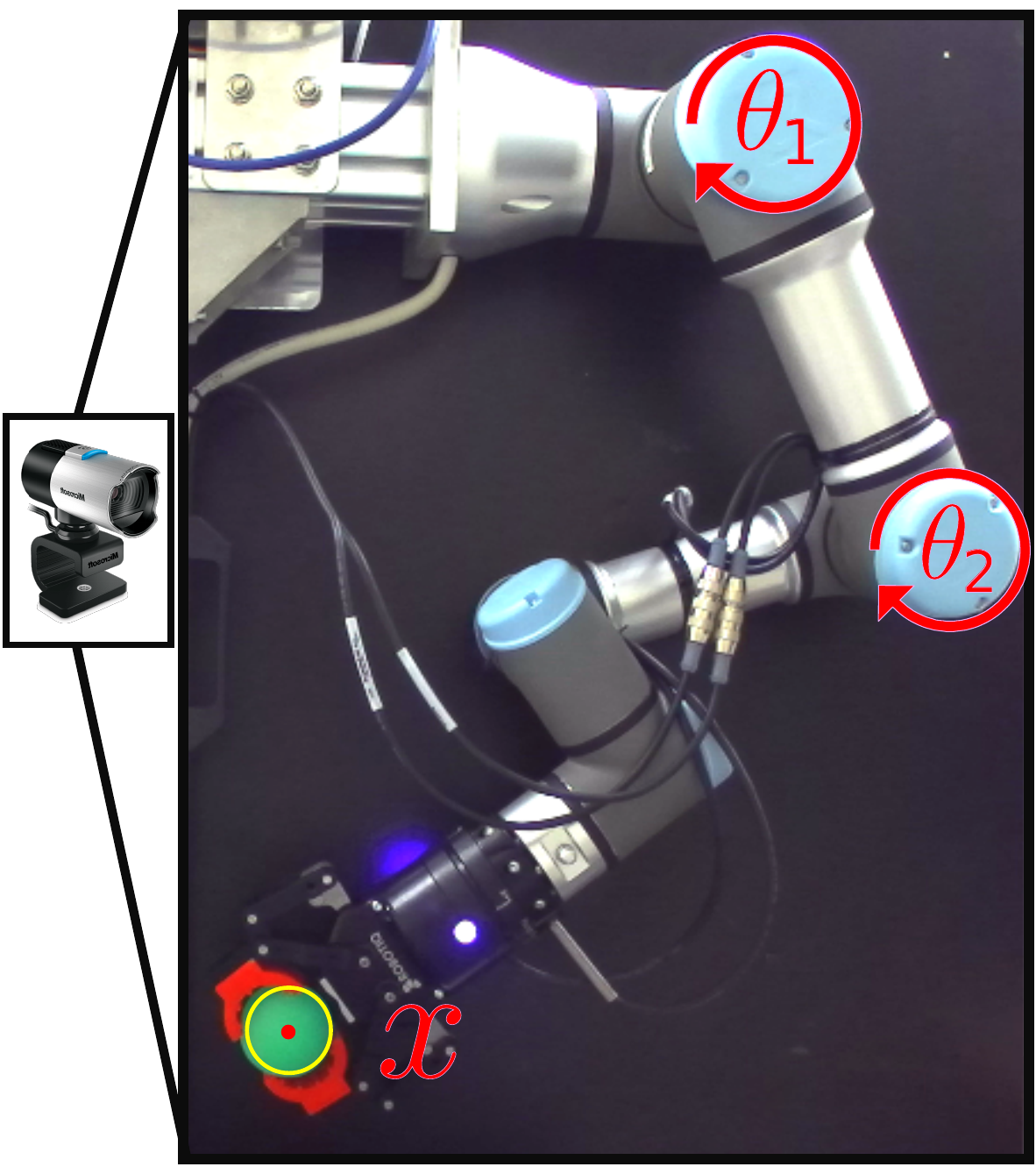}
\caption{}
\label{fig:robot_setup}
\end{subfigure}
\begin{subfigure}{0.49\columnwidth}
\includegraphics[width=\linewidth]{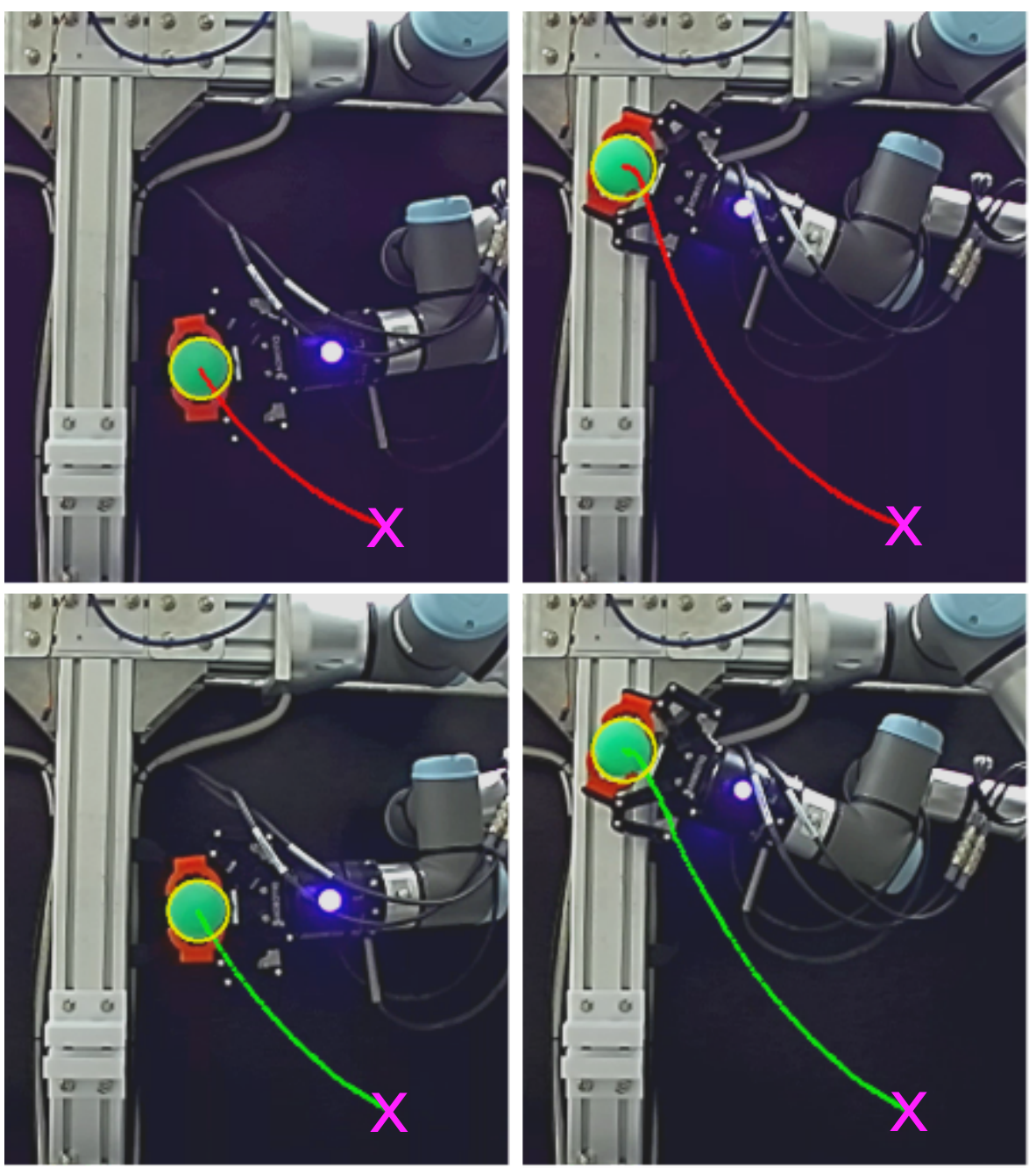}
\caption{}
\label{fig:servo_compare}
\end{subfigure}
\caption{(a) The experimental setup for the UR3 robot; sensory data is received from the camera and robot encoders. (b) The target reaching with (green) and without (red) the cerebellum in action.}
\label{fig:experiments}
\end{figure}

\section{RESULTS}\label{sec:method_ver}
\subsection{Setup}

To test the developed fully spiking control system, the shoulder and elbow of a UR3 robot are controlled to move in a planar workspace, as shown in Fig.\ref{fig:robot_setup}. A camera provides the current position of the end-effector, which is then compared to the target position to generate the desired spatial velocity vector. On the other hand, the motor encoders provide both the angular positions and velocities necessary to represent the joint space. The end-effector is traced by a colored marker with the fixed camera opposing to the relative workspace. The SNN in $DM$ and $CB$ are built by using the NeMo package \cite{nemo} which runs as well on CUDA enabled GPUs (GeForce GTX 1080Ti in our case).

The motor babbling is run to collect the required data for training the network, such that the limits of both the joint space and workspace are defined. The shoulder motion range is [-110\degree,-30\degree] and the elbow motion range is [60\degree,150\degree]. Training the SNN in $DM$ requires around 3000 iterations which are equivalent to 240s in simulation time and 30s in real time, where an iteration means introducing the inputs for 80ms and the output is decoded based on neural activity during that period. Afterwards, the training of the cerebellum model starts, and the learning duration depends on the task to be achieved. Three tasks are conducted as follows.

\begin{figure}[!b]
\centering
\includegraphics[width=\linewidth]{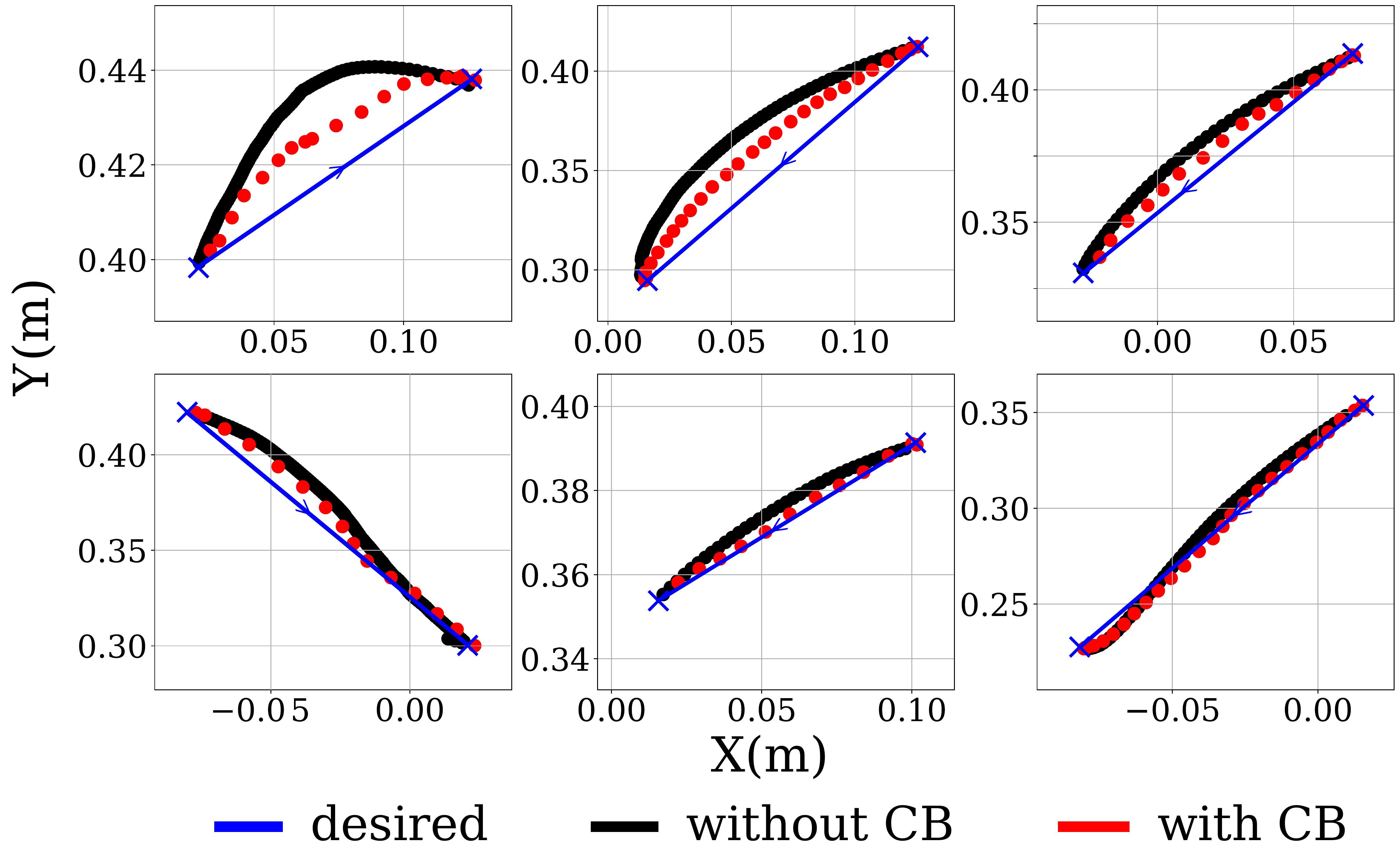}
\caption{Random target reaching.}
\label{fig:rand_servo}
\end{figure}

\begin{figure}[!b]
\centering
\begin{subfigure}{\columnwidth}
\centering
\includegraphics[width=0.9\linewidth]{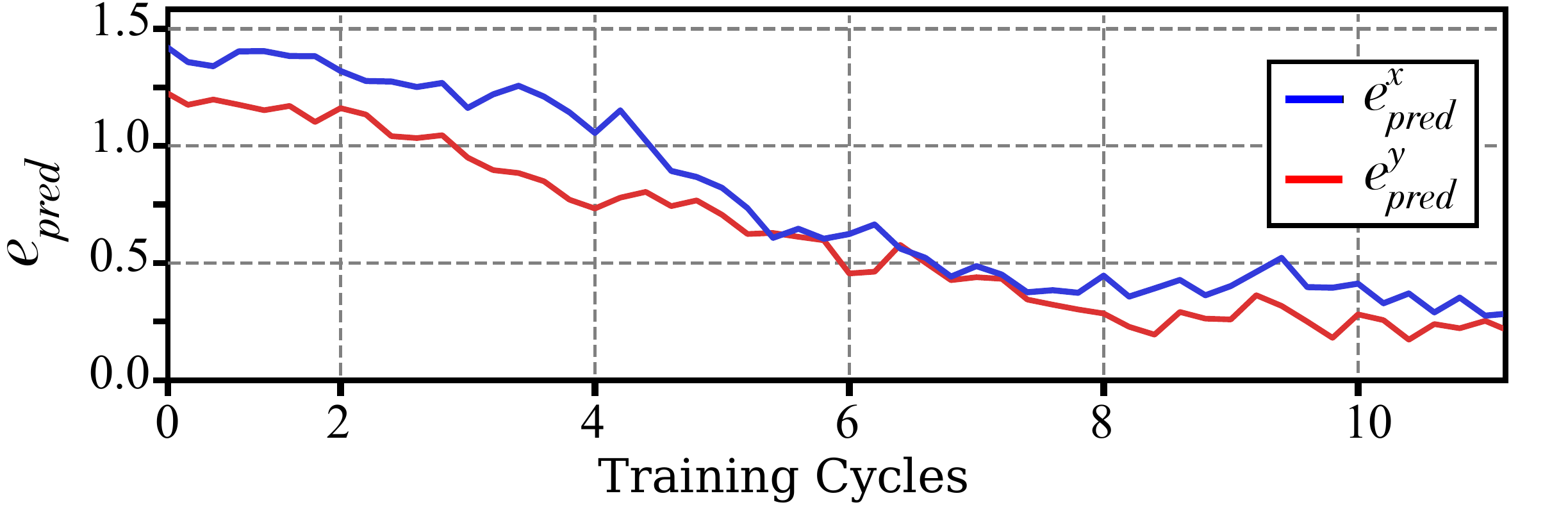}
\caption{}
\label{fig:pred_error}
\end{subfigure}
\centering
\begin{subfigure}{\columnwidth}
\includegraphics[width=0.9\linewidth]{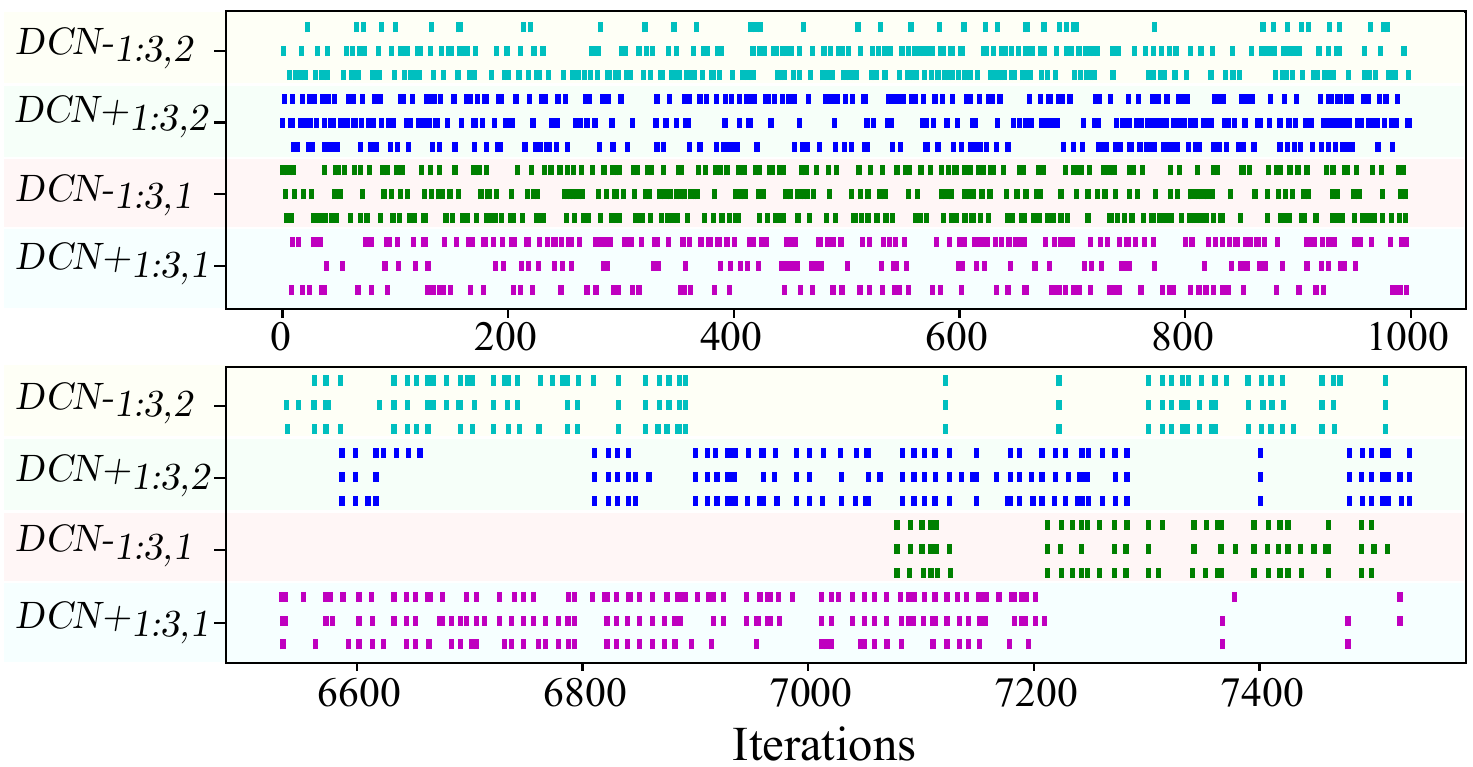}
\caption{}
\label{fig:dcn_firing}
\end{subfigure}
\caption{(a) Discrepancy in predictions of spatial velocity in both $x$ (${e^{x}}_{pred}$) and $y$ (${e^{y}}_{pred}$) directions. A target is chosen randomly, and each training cycle represents a motion attempt from the starting point to the target. (b) Spike raster of neurons in the DCN layer at the beginning and the end of the training for reaching the target.}
\label{fig:rand_trgt}
\end{figure}

\subsection{Random Target Reaching}\label{sec:target_servo}
The robot is instructed to move from/to random target points in the workspace by only relying on the SNN in the $DM$, ie. before the cerebellum model is activated in the control system. After ca. 10000 training iterations, the same motions are repeated with the active cerebellum model, as shown in Fig. \ref{fig:rand_servo}, with an example of the reaching process in Fig. \ref{fig:servo_compare}. The maximum deviation from the desired straight path is reduced with the active cerebellum applied by a mean of 31\%, and a standard deviation of 18\%, while the reaching time is reduced by ca. 75\%. While repeating the reaching for a randomly chosen target (with each reach denoted as a training cycle), the discrepancy between the sensory prediction and the real position of the end-effector is decreased as shown in \ref{fig:pred_error}. When the $PF$ synapses had been modulated, the activity in the $DCN$ changes as well such that the spikes act collectively to encode the sensory predictions as seen in Fig. \ref{fig:dcn_firing}. While these predictions are provided ahead (one iteration earlier), it allows for the modulation of the reference value sent to the $DM$ before attempting the motion. 

\subsection{Radial Reaching}\label{sec:Sequence_servo}

\begin{figure}[!b]
\centering
\begin{subfigure}{0.32\columnwidth}
\centering
\includegraphics[height=0.92\columnwidth]{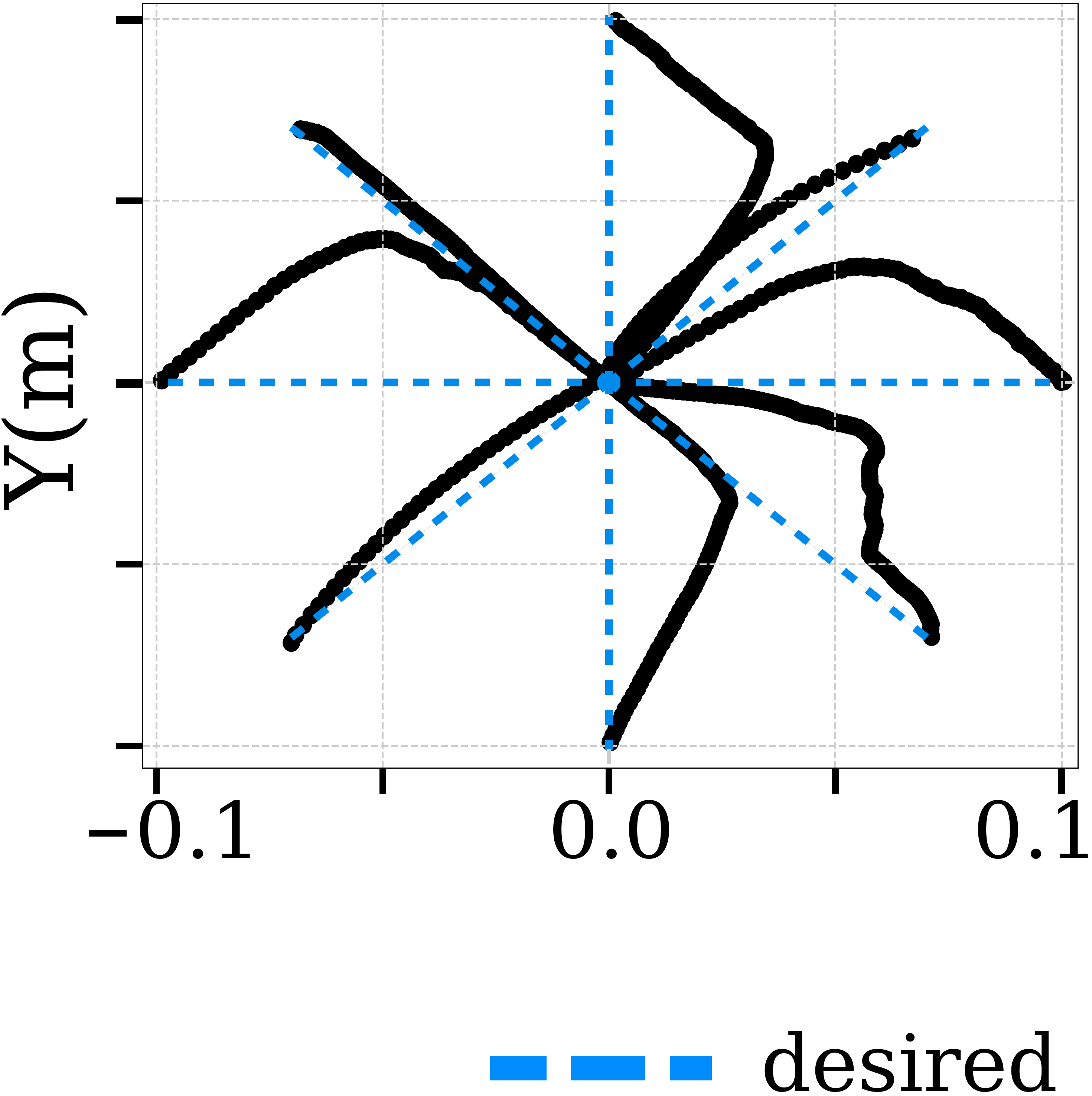}
\caption{}
\label{fig:star_servo_init}
\end{subfigure}
\begin{subfigure}{0.32\columnwidth}
\centering
\includegraphics[height=0.92\columnwidth]{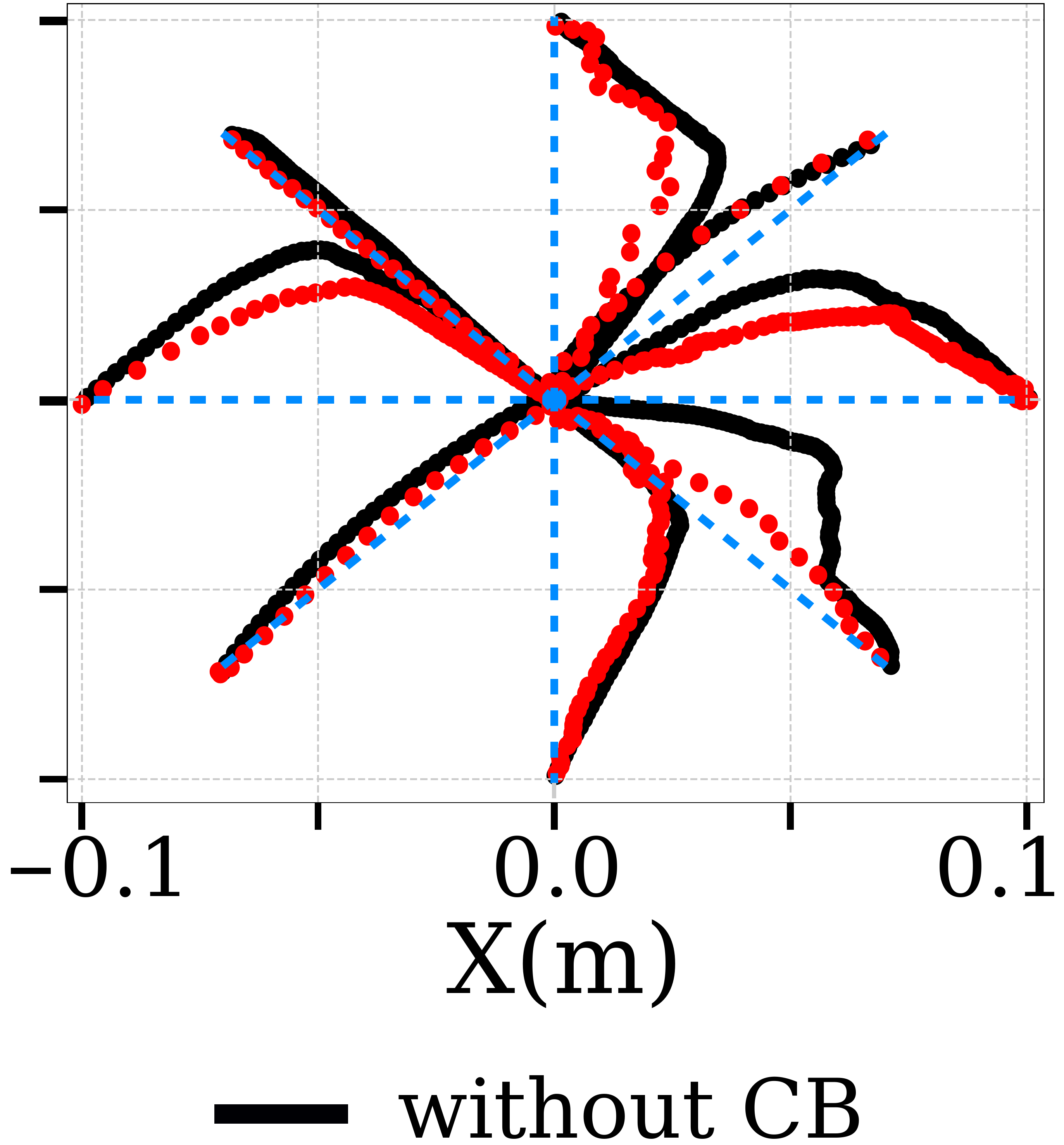} 
\caption{}
\label{fig:star_servo_mid}
\end{subfigure}
\begin{subfigure}{0.32\columnwidth}
\centering
\includegraphics[height=0.92\columnwidth]{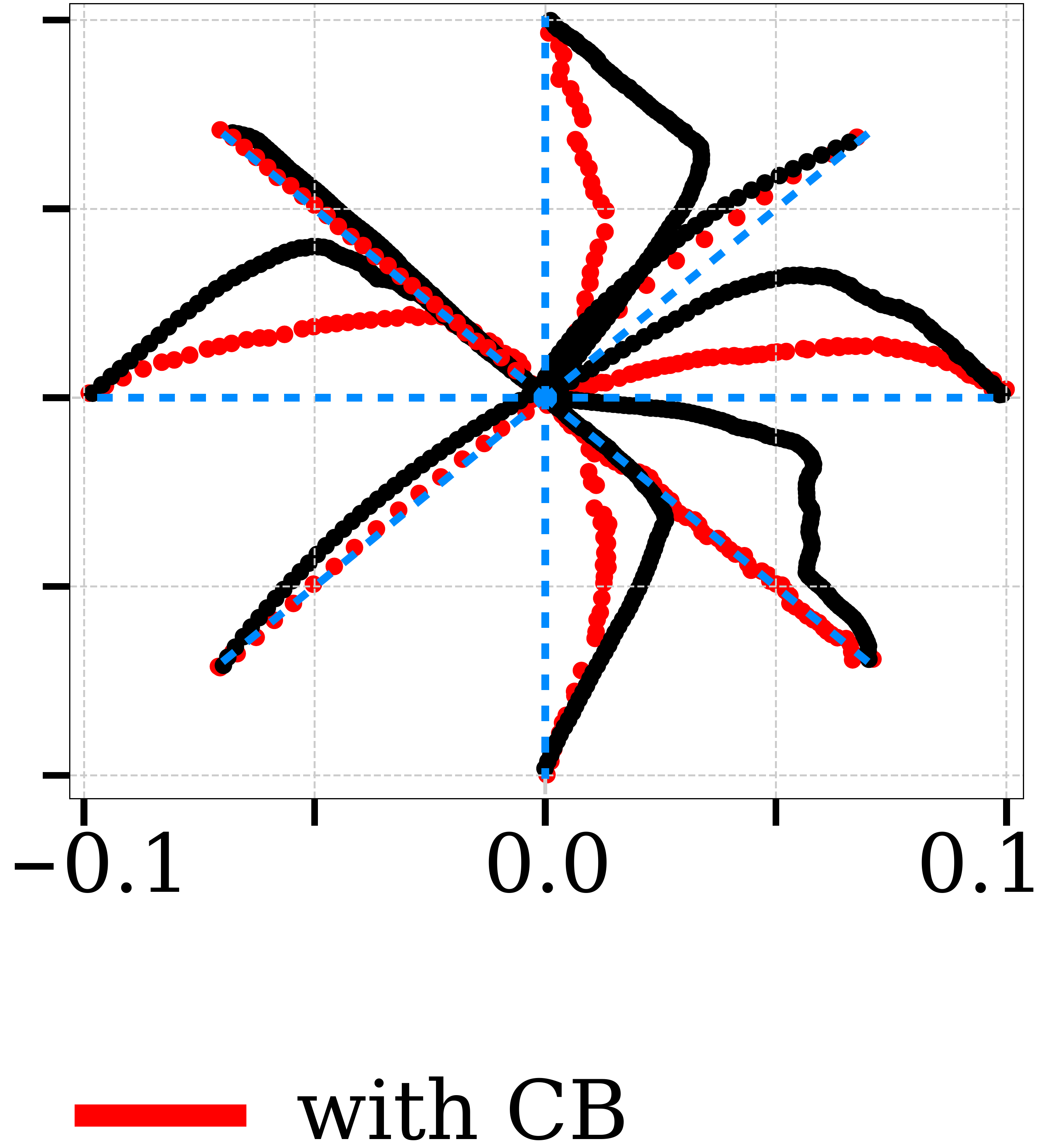} 
\caption{}
\label{fig:star_servo_final}
\end{subfigure}
\caption{The robot moves toward eight target reaching points starting from the same point (0,0) with a 45$\degree$ deviation from each target. The performance is shown in (a) before training the cerebellum; (b) after 9600 iterations of training and 4 repetitions of each target reaching; (c) after 19200 iterations of training and 8 repetitions.}
\label{fig:star_servo}
\end{figure}

\begin{figure}[!b]
\centering
\begin{subfigure}{0.49\columnwidth}
\centering
\includegraphics[width=\columnwidth]{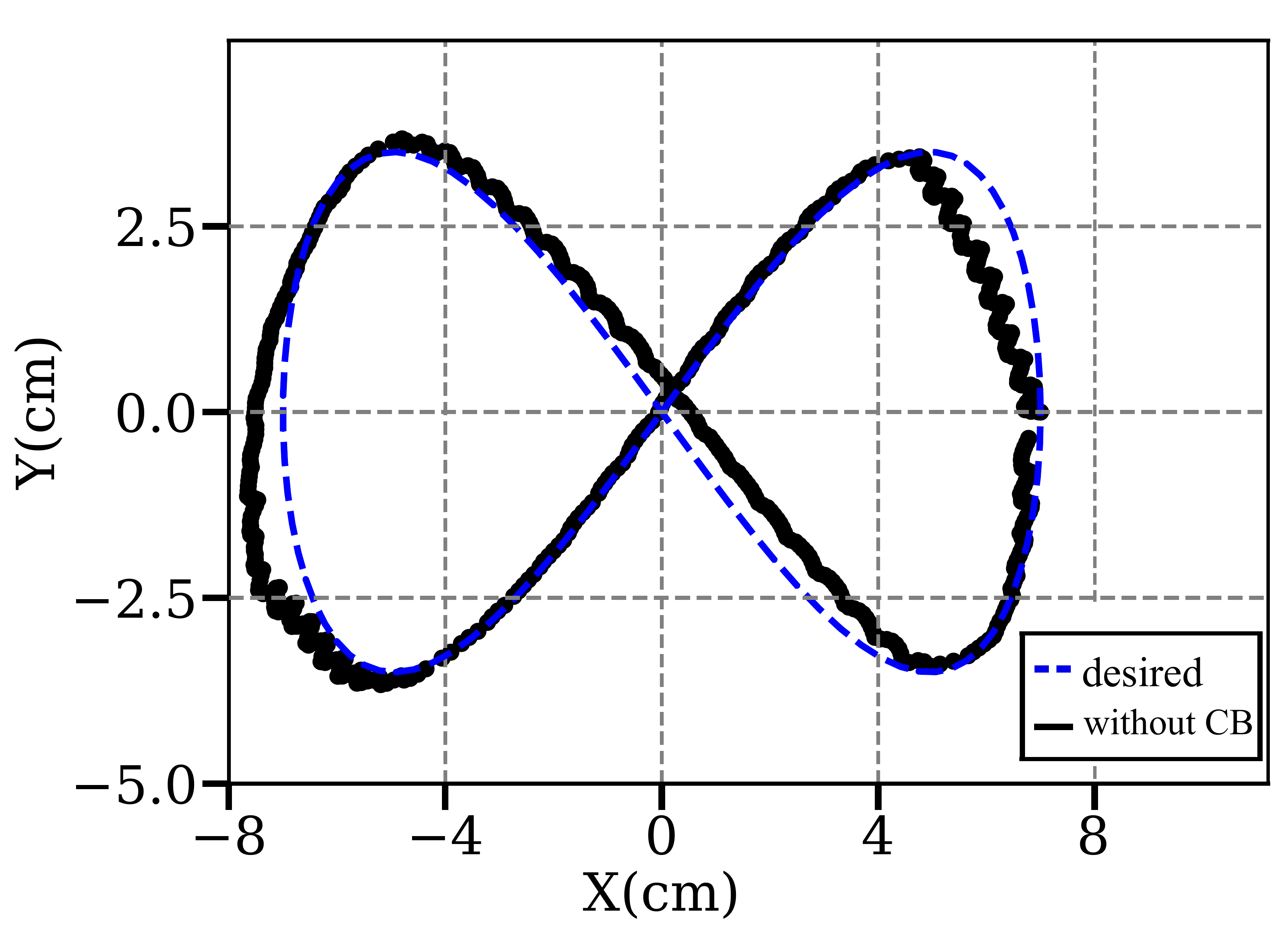}
\caption{}
\label{fig:inf_servo_no_cb}
\end{subfigure}
\begin{subfigure}{0.49\columnwidth}
\centering
\includegraphics[width=\columnwidth]{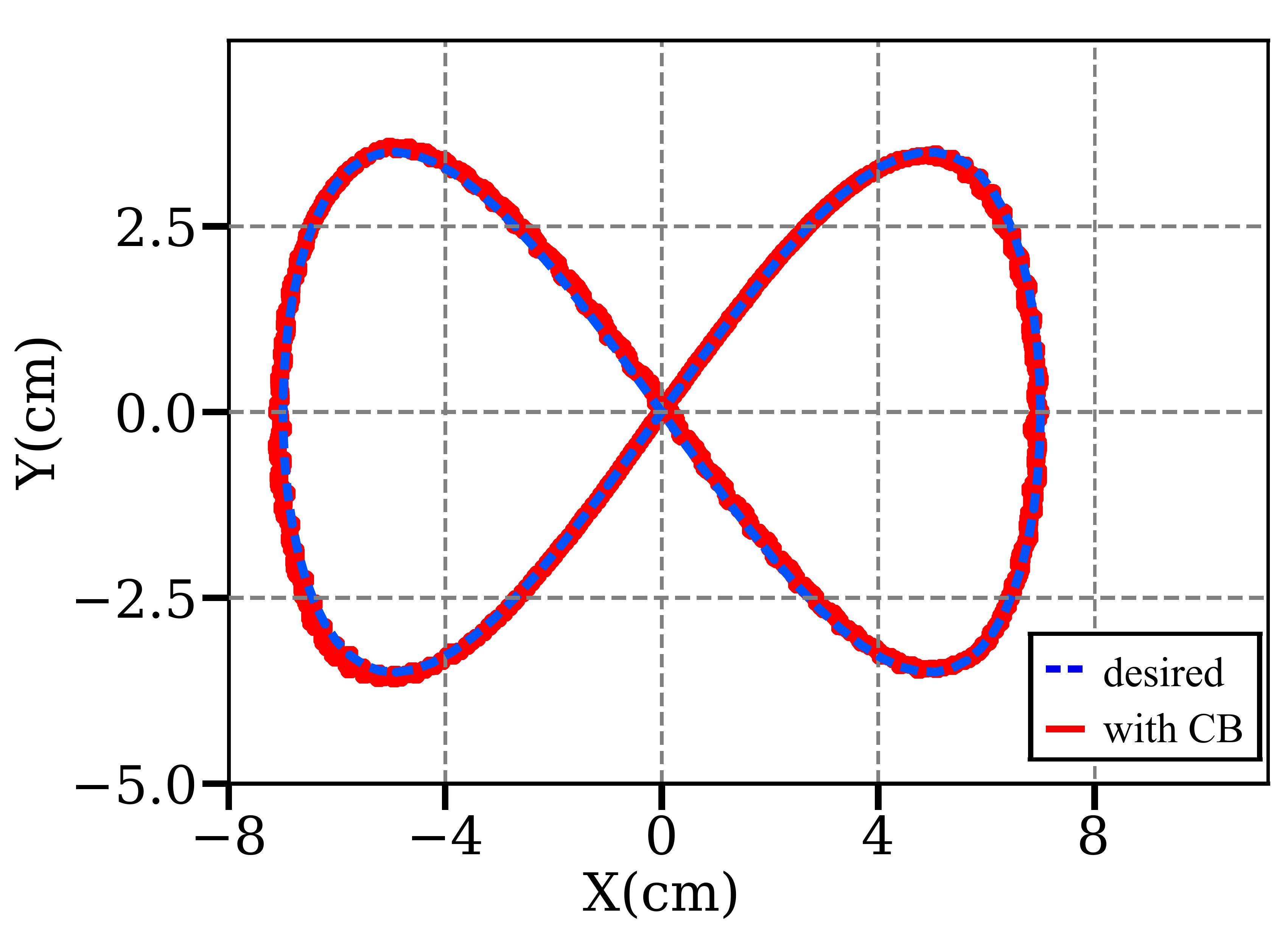}
\caption{}
\label{fig:inf_servo_cb}
\end{subfigure}
\begin{subfigure}{\columnwidth}
\centering
\includegraphics[width=0.8\columnwidth]{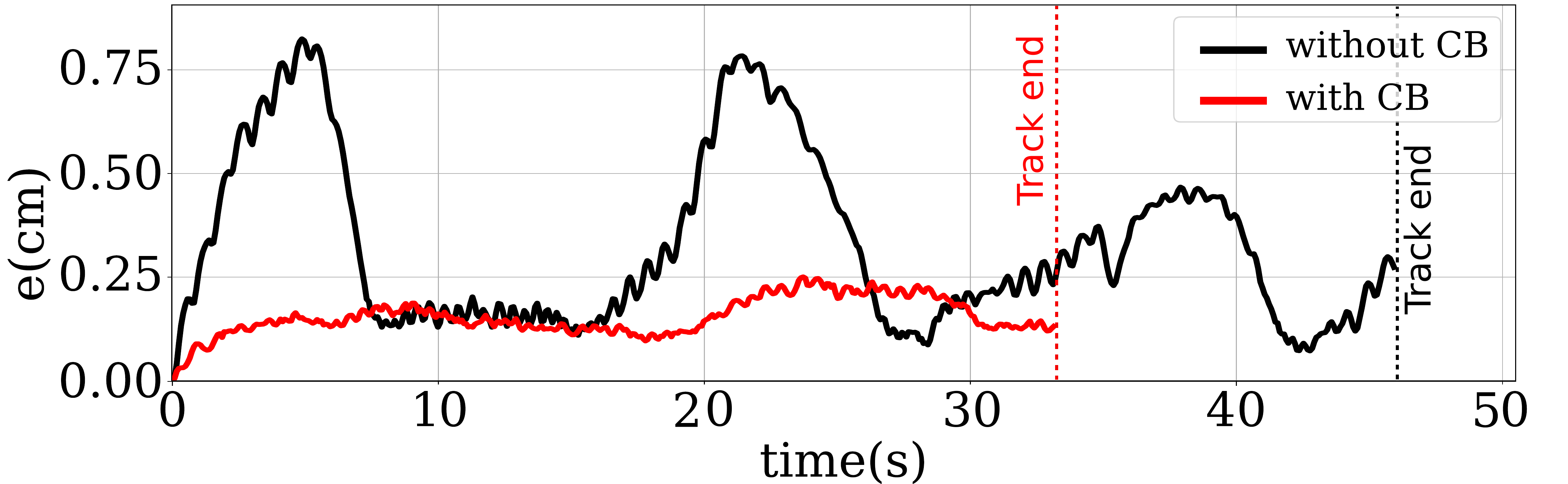}
\caption{}
\label{fig:inf_servo_err}
\end{subfigure}
\caption{Moving the end-effector in an infinity pattern (a) without and (b) with the cerebellum in action. A comparison of deviation (\textit{e}) from the contour and the execution time is plotted in (c).}
\label{fig:infinity_servo}
\end{figure}

The effect of a cerebellar damage was studied in \cite{Fortier2002CEREBELLARAD}, where patients are asked to move at radially arranged targets to check the ability to make smooth and coordinated movements.
Therefore, in this study the robot is instructed to reach target points located in a eight-angled star with an internal angle of 45$\degree$ that are equally 10 cm distant from the central starting point. The robot is required to repeat the reaching motion eight times for each of the 8 targets to give a better performance as shown in Fig. \ref{fig:star_servo}c compared to what obtained after only 4 repetitions (Fig. \ref{fig:star_servo}b) or without the cerebellum in the system (Fig. \ref{fig:star_servo}a).  The maximum deviation is reduced by a mean of 55\% and a standard deviation of
35\%, while the reaching time is reduced by ca. 120\%. It shall be noted that each target is tested independently. The average number of iterations needed to complete a target reach is around 300, which means that approximately 2400 iterations only are needed to train the network to reach in a specific direction compared to 10,000 iterations of random reaching in the previous subsection.

\subsection{Contour Following}\label{sec:track_servo}
The robot end-effector follows a predefined reference contour, an infinitive- or eight-shape ($\infty$) contour (see Fig.\ref{fig:infinity_servo}) defined by $x_{t} = 0.5R_{t}\sin(2\gamma)$ and $y_{t} = R_{t}\cos(\gamma)$, where $R_{t}$ is the radius of each half of the $\infty$-shape (set to 7cm in this experiment), and $\gamma \in [0,2\pi]$ divided over equal intervals to dicretize the contour into a set of points (80 points in this case) as sequential targets. The end-effector shall attempt to reach the point with a defined accuracy (less than 1mm) within a time limit (1$s$), otherwise the target is skipped and the end-effector moves to the next one. The results of the same task carried out by only the $DM$ and with also the $CB$ in action can be seen in Fig.\ref{fig:inf_servo_no_cb} and Fig.\ref{fig:inf_servo_cb}, respectively. The error is defined in this case as the distance between the end-effector position and the closest point on the contour while moving, and it is shown in the plot in Fig.\ref{fig:inf_servo_err} after filtering by a first order filter. It is clear from the plot that the reduction in both the error and time needed to complete the contour following task starting from one point and ending back at the same point. The maximum error is reduced from approximately 8mm to 2.5mm, and the time to complete the contour is reduced from approximately 46$s$ to 33$s$.

\section{CONCLUSIONS}\label{sec:conclusions}
In this study, we suggested a biomimetic control system for vision-based control that integrates a cellular-level spiking cerebellar model that helps to obtain more accurate and faster robot motions. The teaching signal for the cerebellum is based on sensory feedback (in task-space), and thus it allows task-based adaptation in contrast with the previous literature. The sensory corrections are then fed into a motor-cortex-like spiking network which acts as a differential map to convert the motion plan (in spatial coordinates) into motor commands in joint-space. In the developed control system, the cerebellum acts as a Smith predictor to make up for the lack in accuracy in the developed differential map.

From the results, we conclude that both the accuracy (shown as deviation from the desired path) and execution time are reduced thanks to the sensory predictions of the cerebellum. The performance of the robot in reaching a specific target and reaching random targets indicates that the cerebellar-model is able to learn fast (within only 8 repetitions) a specific example and to also generalize its learning to improve the performance across the whole workspace from random motions. Moreover, from the performed experiments, it shall be noted that some regions in the $DM$ have higher error in reaching than other regions. This is mostly related to the randomness in the followed motor babbling technique. Hence, even while providing sensory corrections from the $CB$, the $DM$ still does not hold much improvement at the correct sensory values. This urges the need to apply/develop a different motor babbling algorithm.
The experimental results obtained with the robot arm display a fair correlation with biological observations \cite{Fortier2002CEREBELLARAD, dewolf2016spiking}. Future aim will be to test the effect of cerebellar disfunctions in the control tasks.
Additionally, the current study only considers planar motion of the robot in line with experiments carried out on patients with cerebellar damage. In the future, non-planar motions needs to be studied along with an algorithm for self-tuning the parameters of the cerebellar model. 
While the developed cerebellar model copies just some of the features of the real cerebellum, more biological features are to be included in the future to exploit the full potential of cellular-level models.


\bibliography{biblio.bib}
\bibliographystyle{IEEEtran}

\end{document}